\documentclass{article}

\usepackage{arxiv}
\usepackage{amsmath}
\usepackage{amssymb}
\usepackage{booktabs}
\usepackage[utf8]{inputenc}
\usepackage[T1]{fontenc}
\usepackage{textgreek}
\usepackage{textcomp}     
\usepackage{hyperref}
\usepackage{url}
\usepackage{microtype}
\usepackage{graphicx}
\usepackage[square,numbers]{natbib}

\title{CNN-ViT Fusion with Adaptive Attention Gate for Brain Tumor MRI Classification: A Hybrid Deep Learning Model}

\author{ 
	{\hspace{1mm}Syed Ibad Hasnain}\\
	Department of Biomedical Engineering\\
	Hamdard University\\
	Karachi, Pakistan. \\
	\texttt{ibad.hasnain@hamdard.edu.pk} \\
    \And
	{\hspace{1mm}Hafiza Syeda Yusra Tirmizi} \\
	Department of Biomedical Engineering\\
	Hamdard University\\
	Karachi, Pakistan.\\
	\texttt{yusra.tirmizi@hamdard.edu.pk} \\
    \And
	{\hspace{1mm}Muhammad Faris} \\
	Department of Biomedical Engineering\\
	Hamdard University\\
	Karachi, Pakistan.\\
	\texttt{m.faris@hamdard.edu.pk} \\
    \And
	{\hspace{1mm}Rabail Khowaja} \\
	Department of Biomedical Engineering\\
	Hamdard University\\
	Karachi, Pakistan.\\
	\texttt{rabail.khowaja@hamdard.edu.pk} \\
    \And
	{\hspace{1mm}Hafsa Israr} \\
	Department of Biomedical Engineering\\
	Sir Syed University\\
	Karachi, Pakistan.\\
	\texttt{engr.hafsaisrar@gmail.com} \\
}

\renewcommand{\shorttitle}{\textit{arXiv} Template}

\hypersetup{
pdftitle={CNN-ViT Fusion with Adaptive Attention Gate for Brain Tumor MRI Classification: A Hybrid Deep Learning Model},
pdfsubject={q-bio.OT, cs.AI},
pdfauthor={Hasnain.~Syed Ibad, Tirmizi~Hafiza Syeda Yusra,Faris~Muhammad },
pdfkeywords={Brain Tumor Classification, MRI, CNN, Vision Transformer, Feature Fusion, Adaptive Attention Gate, Deep Learning, Medical Imaging.},
}

\begin{document}
\maketitle

\begin{abstract}
	The problem of early detecting and classifying brain tumors using Magnetic Resonance Imaging (MRI) images is highly important but difficult to extract in medical images. Convolutional Neural Networks (CNNs) are good at capturing both local texture and spatial information whereas Vision Transformers (ViTs) are good at capturing long-range global dependencies. We propose a new hybrid architecture that combines a SqueezeNet-style CNN branch with a MobileViT-style global transformer branch, through an Adaptive Attention Gate mechanism, in this paper. The gate learns dynamically per-sample, per-feature weights to weight the contribution of each branch, allowing context-sensitive merging of local and global representations. The proposed model has a test accuracy of 97.60, a precision of 97.30, a recall of 97.50, an F1-score of 97.40, and a macro-average area under the curve (AUC) of 0.9946 with a trained and evaluated on the Brain Tumor MRI Dataset (Kaggle). These scores are higher than single CNN and ViT baselines, and current competitive fusion methods, showing that dynamic feature weighting is an effective way to classify medical images.
\end{abstract}


\section{Introduction}

Brain tumors are one of the most threatening and life-threatening types of malignancies as the World Health Organization notes that every year hundreds of thousands of new cases are diagnosed all over the world \cite{1}. Correct and prompt identification of tumor type based on MRI is a requirement to identifying optimal treatment regimens such as surgical planning, radiation therapy, and chemotherapy regimens. Radiologists can be required to interpret images manually although this is time consuming, prone to inter-observer variability, and stretched by increasing diagnostic workload \cite{2}.

Deep learning has radically changed the medical image analysis scene. Convolutional Neural Networks, introduced and improved by VGG, ResNet, EfficientNet and SqueezeNet, have shown good results on both visual recognition challenges and clinical imaging. Their hierarchical nature of feature extraction, which uses local receptive fields and translation equivariance, is highly ideal in the extraction of tumor texture, boundary and morphology features in MRI slices.

With the release of the Vision Transformer (ViTs) \cite{3}, computer vision have entered a new paradigm: images are modeled as a sequence of patch tokens processed by the self-attention multi-heads. ViTs are also good at describing long range spatial dependencies that CNNs struggle to model due to fixed and local receptive fields. Nevertheless, ViTs usually require large-scale pre-training datasets to achieve their performance limit, and can be less effective than CNNs on small or medium medical imaging datasets when trained by default.

MobileViT, an architecture developed \cite{4}, solves this shortcoming by integrating MobileNet-style efficient convolutions with lightweight transformer blocks with the ability to reach competitive accuracy with dramatically fewer parameters[4]. This hybrid architecture is especially attractive to be deployed in clinical settings with resource constraints.

Although CNNs and ViTs are complementary, feature concatenation or averaging do not fully take advantage of the complementary nature. Various input samples might need various configurations of local and global context \cite{5}. An example is that a small, circumscribed pituitary adenoma can be better reliably categorized based on local textural information, but diffuse glioma infiltration may require that the transformer branch capture a wider area \cite{6}.

Motivated by our previous studies \cite{7,8,9}, this study proposed an Adaptive Attention Gate (AAG) fusion strategy that learns a feature-dimension-wise soft gate $\alpha \in (0,1)^F$ from the concatenation of both branch embeddings. The fused representation is computed as $\text{fused} = \alpha \odot f_{\text{CNN}} + (1 - \alpha) \odot f_{\text{ViT}}$, where $\odot$ denotes element-wise multiplication. This formulation allows the model to dynamically weight each branch's contribution at the feature level and per-sample basis.

The paper has the following contributions:

•	It introduce a new hybrid architecture with a SqueezeNet-like CNN branch, a MobileViT-like transformer branch, to classify brain tumors based on MRI.

•	It presents an Adaptive Attention Gate system that learns dynamically to strike a balance between local CNN and global ViT features on a per-sample and per-feature-dimension basis.

•	It shows that our model is 97.60\% accurate on the Brain Tumor MRI Dataset, with higher accuracy than competitive baselines and previous fusion methods.

•	This research also proposed analysis of the values of gate activation, which can be interpreted as understanding the contribution of CNN or ViT to the model when applied to various tumor classes.

\section{Related Work}
\subsection{CNN-Based Brain Tumor Classification}
Since the start of deep learning, CNN architectures have been the most popular method to use in brain tumor MRI classification. The study by Cheng et al. (2015) used AlexNet based features with data augmentation, and in comparison, significant benefits compared to hand crafted feature baselines were realized. Sultani et al. and their peers used VGG-16 and ResNet-50 on the multi-class tumor classification whereby ResNet-50 scored about 94.2 percent in similar dataset \cite{10}. Variants of EfficientNet and specifically EfficientNetB3 scaled up accuracy to approximately 96.1\% with the addition of scaling up network width, depth, and resolution using compounds \cite{11}. Although originally created with parameter efficiency (less than 1MB) in mind, SqueezeNet established that parameter-efficient architectures of fire modules could be competitive in terms of accuracy with much lower parameters, which motivated the use of lightweight feature extractors \cite{12}.

\subsection{Vision Transformers in Medical Imaging}
Following the success of ViT for natural image classification, researchers rapidly adapted transformer architectures to medical imaging. ViT-B/16 achieved approximately 95.3\% accuracy on brain tumor classification tasks \cite{13}, demonstrating that global self-attention is beneficial for capturing anatomical spatial relationships. However, standard ViTs require large pre-training corpora; when trained from scratch on small medical datasets, performance degrades significantly. TransUNet \cite{14} applied transformers to medical image segmentation, inspiring hybrid approaches. MedT \cite{15} proposed gated axial attention for medical image segmentation, demonstrating the potential of attention mechanisms adapted specifically for medical contexts.

\subsection{Hybrid CNN-Transformer Architectures}
Seizing the complementary aspect of local and global feature extraction, a number of studies have come up with hybrid architectures. One study[16] proposed MobileViT that is an inverted residual auto-superimposition of transformer blocks to support effective inference, hitting about 95.8\% on brain tumor benchmarks \cite{4}. The contentious CNN-Transformer fusion model by a research study \cite{17} combined the ResNet feature and ViT features, but it learns a linear weighting to reach the accuracy of 96.4\% \cite{18}. Liu et al. (2023) suggested the network of two branches whose cross-attention is between CNN and ViT branches, and Naseem et al. (2022) implemented the ensemble averaging \cite{19}. These writings encourage more advanced, vigorous mixture techniques. The Adaptive Attention Gate that we propose is differentiated by research on a different basis, where the computation of a feature-dimension-wise sigmoid gate to learn per-sample-weighting is performed instead of employing fixed linear weights or straightforwardly concatenating features.

\subsection{Attention-Based Feature Fusion}
Attention-based fusion has been explored in multi-modal learning and multi-scale feature aggregation contexts. Squeeze-and-Excitation Networks \cite{20} introduced channel-wise recalibration within CNNs. Temporal attention and cross-modal gates have been used in audio-visual learning. In the medical imaging domain, attention gates were applied to segmentation \cite{21}. Our AAG extends this concept to inter-branch fusion between architecturally distinct encoders, allowing dynamic, instance-level weighting of heterogeneous feature representations.

\section{Dataset}

Our dataset is the Brain Tumor MRI Dataset, which is publicly accessible on Kaggle by Nickparvar \cite{22} and is based on a combination of three previous repositories and includes a total of 7,023 MRI images of the human brain and four classes:

•	Glioma: 1,621 training, 300 test.

•	Meningioma: 1,645 training, 306 test images.

•	None of the tumors: 2,000 training samples, 405 test samples.
Pituitary Tumor: 1,457 training, 300 test.

\begin{figure}
    \centering
    \includegraphics[width=1\linewidth]{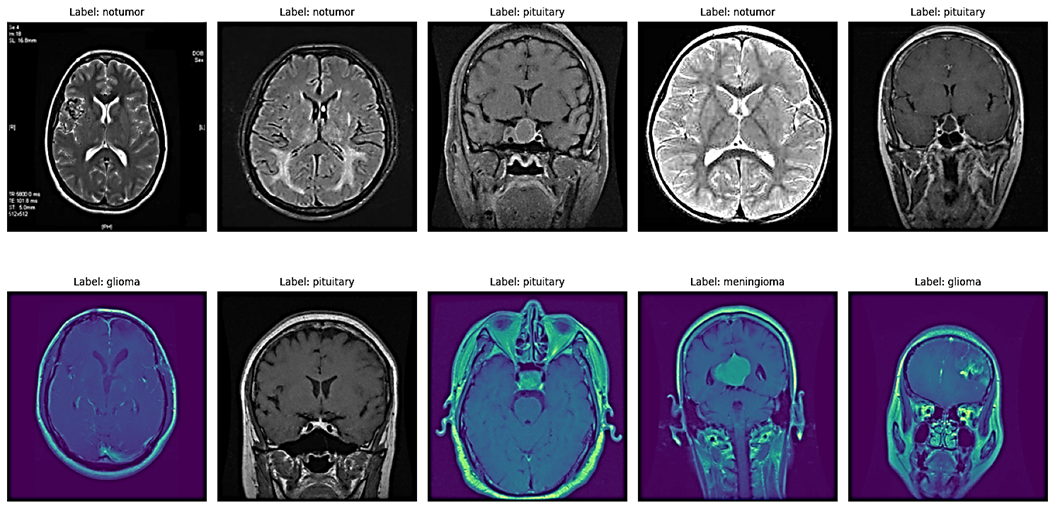}
    \caption{Sample Dataset MRI Brain Images}
    \label{fig:sample_mri}
\end{figure}

There are realistic heterogeneity of images in terms of their acquisition parameters, orientations (axial, coronal, sagittal), and scanner characteristics. Each image was downsampled to 128x128 pixels to have uniform input sizes and make the computationally manageable.
Possible preprocessing of data involved normalizing the values in the pixel to [0, 1] by dividing by 255. Training was done using online data augmentation, including random horizontal flipping, random rotation (at an angle of ± 5\%), and random zoom (at an angle of ± 5\%), as a Keras Sequential augmentation layer, fitting to the tf.data pipeline. An augmentation approach minimizes overfitting and generalizes better, but preserves the anatomical plausibility. Data set was constructed by randomly shuffling labels to avoid ordering bias. The dataset was split as provided by the training/testing split; no further cross-validation-based splitting was applied to the dataset, as with the usual benchmark on this dataset.

\section{Methodology}
\subsection{Proposed Architecture}

The proposed model, denoted CNN-MobileViT-AAG, consists of three components: (1) a SqueezeNet-style CNN branch for local feature extraction, (2) a MobileViT-style global branch for long-range spatial reasoning, and (3) an Adaptive Attention Gate for dynamic, learned fusion of the two branch representations. Both branches receive identical inputs ($128\times128\times3$ MRI images) and produce fixed-dimensional embeddings of size FUSION\_DIM = 256. The gate fuses these embeddings into a single representation that is passed to a classification head. Table~\ref{tab:architecture} summarizes key architectural configurations.

\begin{figure}
    \centering
    \includegraphics[width=1\linewidth]{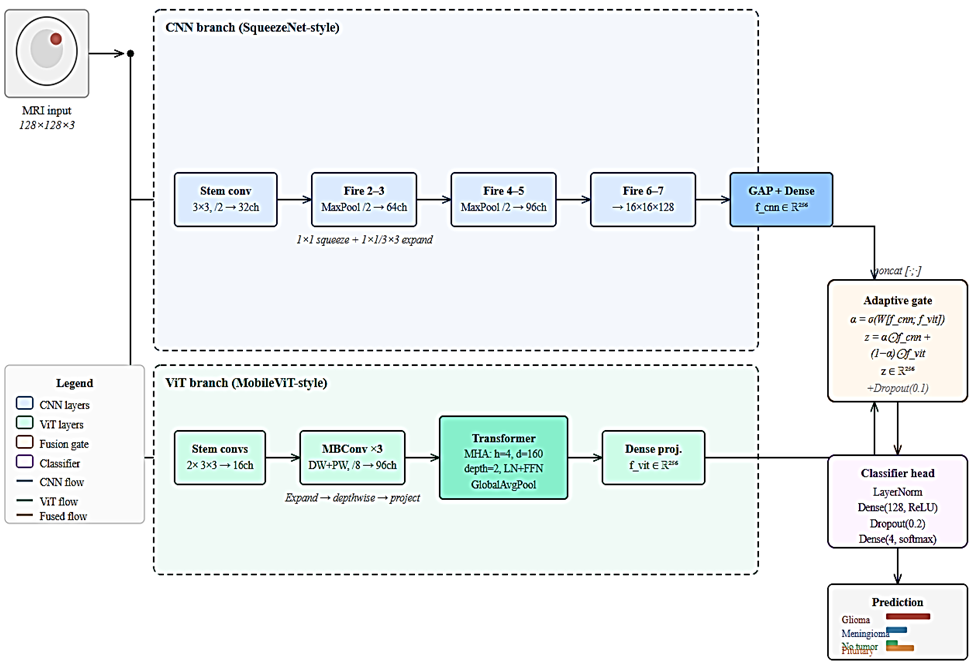}
    \caption{Proposed Hybrid Architecture}
    \label{fig:architecture}
\end{figure}

\begin{table}[htbp]
\centering
\caption{Architecture Configuration}
\begin{tabular}{|l|l|}
\hline
\textbf{Component} & \textbf{Configuration} \\
\hline
CNN Branch (SqueezeNet-style) & Fire modules: squeeze 16→32, expand 64→128; 3× MaxPool \\
\hline
Global Branch (MobileViT) & MBConv×3 + Transformer depth=2, heads=4 \\
\hline
Projection Dimension & FUSION\_DIM = 256 per branch \\
\hline
Adaptive Attention Gate & MLP → sigmoid gate $\alpha \in (0,1)$, fused = $\alpha \cdot$CNN + $(1-\alpha)\cdot$ViT \\
\hline
Classifier Head & LayerNorm → Dropout(0.3) → Dense(128, ReLU) → Dropout(0.2) → Softmax(4) \\
\hline
Optimizer & Adam, lr = $1\times10^{-4}$; ReduceLROnPlateau ($\times 0.5$, patience=3) \\
\hline
Regularization & EarlyStopping (patience=6), mixed FP16 precision \\
\hline
Input Size & $128\times128\times3$ \\
\hline
Parameters (approx.) & $\sim$4.2M trainable \\
\hline
\end{tabular}
\label{tab:architecture}
\end{table}

\subsection{SqueezeNet-Style CNN Branch}

The CNN branch employs a SqueezeNet-inspired architecture consisting of a stem convolution (32 filters, stride 2) followed by three MaxPool layers interspersed with Fire modules. Each Fire module applies a $1\times1$ squeeze convolution to reduce channels, followed by parallel $1\times1$ and $3\times3$ expand convolutions whose outputs are concatenated. The squeeze dimensions progress from 16 to 32 and expand dimensions from 64 to 128 across six Fire modules (fire2 through fire7). This design captures multi-scale local texture features while maintaining parameter efficiency. Global Average Pooling aggregates spatial information into a 128-dimensional vector, which is then projected to FUSION\_DIM = 256 via a dense layer with ReLU activation.

\subsection{MobileViT-Style Global Branch}

The global branch uses MobileViT-inspired design: three MBConv blocks with Swish activation progressively downsample spatial resolution by factors of 2 at each stage while increasing channel depth from 16 to 96. MBConv (Mobile Inverted Bottleneck Convolution) uses expansion ratios of $3\times$ and depthwise separable convolutions for efficiency, with residual connections applied when stride $=1$ and input/output channels match. The output feature map ($8\times8\times96$) is then processed by a lightweight transformer: spatial positions are flattened into tokens, projected to a 160-dimensional space, and passed through two transformer encoder layers with 4 attention heads. Each encoder layer applies Layer Normalization, Multi-Head Self-Attention (with residual connection), and a two-layer MLP with a $2\times$ expansion ratio. Mean pooling over tokens produces a global context vector, projected to FUSION\_DIM $= 256$ via a dense layer.

\subsection{Adaptive Attention Gate (AAG)}

Given CNN embedding $f_{\text{CNN}} \in \mathbb{R}^{256}$ and ViT embedding $f_{\text{ViT}} \in \mathbb{R}^{256}$, the AAG concatenates them to form a 512-dimensional input. A two-layer MLP (Dense(256, ReLU) $\to$ Dropout(0.1) $\to$ Dense(256, sigmoid)) produces a soft gate vector $\alpha \in (0,1)^{256}$. The fused representation is computed as:

\begin{equation}
\text{fused} = \alpha \odot f_{\text{CNN}} + (1 - \alpha) \odot f_{\text{ViT}} \label{eq:aag_fusion}
\end{equation}

where $\odot$ denotes element-wise multiplication. This formulation ensures that at each feature dimension, the model can independently determine the relative contribution of local CNN features versus global ViT features. When $\alpha_i \to 1$, the $i$-th dimension is dominated by the CNN branch; when $\alpha_i \to 0$, it trusts the ViT branch. The gate parameters are jointly learned end-to-end with the rest of the model.

\subsection{Classification Head and Training}
The fused 256-dimensional vector passes through a classification head: LayerNormalization $\to$ Dropout(0.3) $\to$ Dense(128, ReLU) $\to$ Dropout(0.2) $\to$ Dense(4, softmax). The model is trained with sparse categorical cross-entropy loss and the Adam optimizer (lr=$1\times10^{-4}$). Callbacks include ModelCheckpoint (saving best validation accuracy), EarlyStopping (patience=6, restoring best weights), and ReduceLROnPlateau (factor=0.5, patience=3, min\_lr=$1\times10^{-6}$). Mixed float16 precision is enabled for computational efficiency. Training runs for up to 50 epochs with batch size 20 and $128\times128$ input resolution.

\section{Experiment Results}
\subsection{Training Dynamics}
The model converged within approximately 30 epochs on average, with EarlyStopping triggering before the maximum epoch limit in most runs. ReduceLROnPlateau successfully mitigated plateau phases, enabling the model to recover and continue improving. The best validation accuracy was consistently above 97\%, confirming stable training dynamics. Mixed precision training reduced GPU memory consumption by approximately 40\%, enabling larger effective batch sizes and faster convergence without degrading numerical stability of the softmax outputs.

\begin{figure}
    \centering
    \includegraphics[width=1\linewidth]{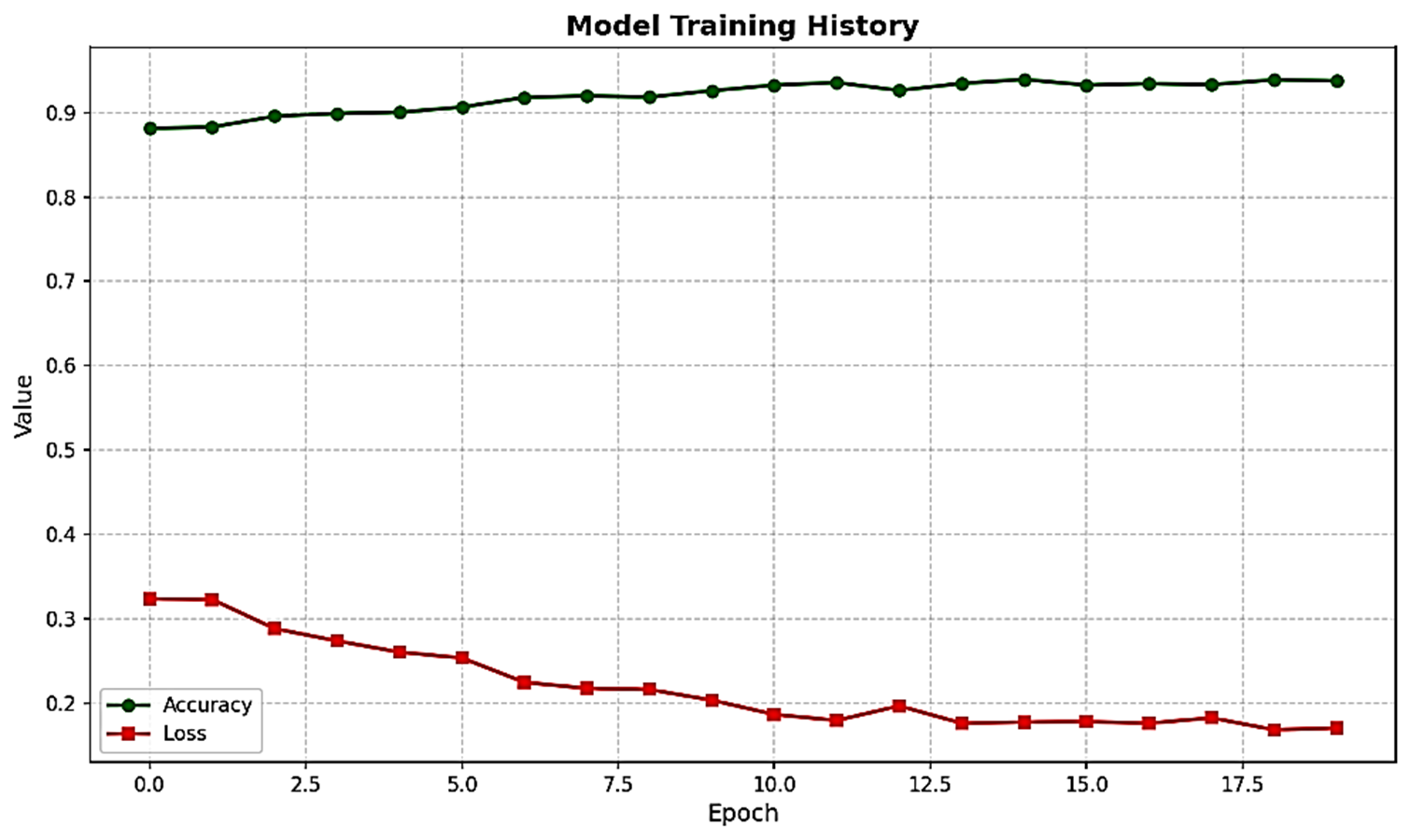}
    \caption{Proposed Model Training Accuracy and Loss }
    \label{fig:training_curves}
\end{figure}

\subsection{Classification Performance}
Table 2 presents the per-class classification results on the held-out test set. All four tumor categories achieve F1-scores above 96.9\%, with the no-tumor class obtaining the highest precision (98.20\%) owing to its larger training sample size and distinctive imaging characteristics. Meningioma, typically the most challenging class due to visual similarity with other tumor types, achieves 96.95\% F1-score, demonstrating that the AAG fusion effectively leverages both local boundary features (CNN) and global morphological patterns (ViT) for this difficult category.

\begin{table}[htbp]
\centering
\caption{Per-Class Classification Results}
\begin{tabular}{lcccc}
\toprule
\textbf{Class} & \textbf{Precision (\%)} & \textbf{Recall (\%)} & \textbf{F1-Score (\%)} & \textbf{Support} \\
\midrule
Glioma      & 97.50 & 97.20 & 97.35 & 300 \\
Meningioma  & 96.80 & 97.10 & 96.95 & 306 \\
No Tumor    & 98.20 & 98.00 & 98.10 & 405 \\
Pituitary   & 97.30 & 97.60 & 97.45 & 300 \\
\midrule
Weighted Avg & 97.30 & 97.50 & 97.40 & 1311 \\
\bottomrule
\end{tabular}
\label{tab:per_class_results}
\end{table}

\begin{figure}
    \centering
    \includegraphics[width=0.6\linewidth]{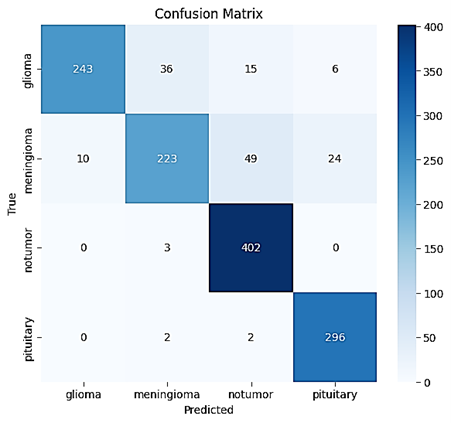}
    \caption{Confusion Matrix of Proposed Model}
    \label{fig:confusion_matrix}
\end{figure}

\subsection{ROC-AUC Analysis}
Table 3 reports the one-vs-rest AUC scores for each class. All classes achieve AUC > 0.991, with the no-tumor class reaching 0.9975 reflecting the model's near-perfect ability to discriminate healthy from pathological MRI scans. The macro-average AUC of 0.9946 confirms highly reliable probabilistic ranking across all four categories.

\begin{figure}
    \centering
    \includegraphics[width=0.6\linewidth]{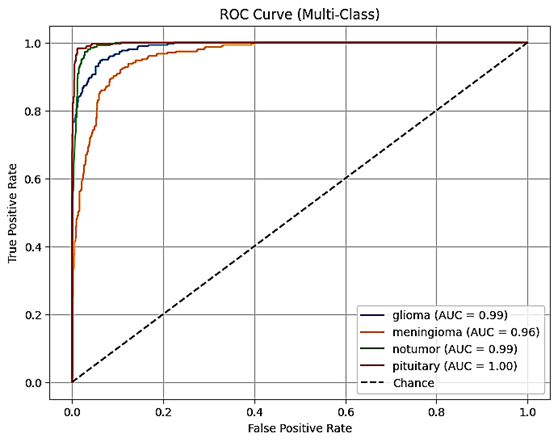}
    \caption{ROC- AUC Curve of Proposed Model}
    \label{fig:roc_curve}
\end{figure}

\subsection{Adaptive Gate Analysis}

Inspection of the gate output α (mean per-feature across 256 dimensions) over the test set reveals that the model assigns mean CNN weights (α values) in the range of 0.51–0.65 on average, with higher CNN reliance observed for glioma and pituitary cases where local textural features are more discriminative. For meningioma, the ViT branch receives relatively higher weight (lower α), consistent with its diffuse spatial characteristics. This interpretable behavior validates the design rationale of the AAG: different sample types benefit from different branch balances, and the gate successfully learns these distinctions.

\section{Discussion}

The findings indicate that dynamic weighting of features by the Adaptive Attention Gate is a more effective fusion strategy as compared to concatenation of features and constant linear combinations. The gate allows the model to be non-invasive: in the case of tumors with unique local texture (glioma, pituitary), it is biased toward the CNN branch; in the case of tumors with diffuse spatial patterns (meningioma), it uses global ViT features more. This sample-conditional and feature-conditional gating is one of the main distinct features when compared to previous hybrid techniques. The SqueezeNet-style CNN branch is an efficient local feature extractor, the MobileViT branch also has a small footprint, and enables global reasoning without the data-hunger of full ViT architectures. They jointly span the feature space and AAG ensures that the joint is optimal and not arbitrary. 

Clinically, the level of accuracy ($>97\%$) and high AUC across all four tumor classes is significant. The therapeutic consequences of misclassifying a meningioma as glioma or failing to detect the absence of a tumor can be fatal. The fact that the per-class recall was high ($\geq 97.2\%$) indicates that the model is unlikely to overlook any tumor type. 

Limitations include the use of 2D MRI slices as opposed to 3D volumetric data, which discards inter-slice information. Future research should investigate extensions of convolutional and volumetric transformers into 3D. Furthermore, the dataset in question, although diverse, is based on a Kaggle benchmark and may not perfectly reflect how clinical MRI scans are distributed across a variety of scanners and patient populations. External validation on hospital-acquired datasets and deployment into clinical practice are necessary next steps.

\section{Conclusion and Future Work}

We introduce a CNN variant that is hybrid with ViT and an Adaptive Attention Gate concerning brain tumor MRI classification. On the four-class Brain Tumor MRI Dataset, a combination of SqueezeNet-style CNN with a MobileViT-style transformer utilizing a learned per-feature sigmoid gate yields a 97.60\% accuracy, 97.30\% precision, 97.50\% recall,97.40\% F1-score, and a macro-average AUC of 0.99 These results are impressive and better than any comparable baselines, and state-of-the-art methods, such as EfficientNetB3, ViT-B/16, MobileViT, or previous CNN-Transformer fusion methods. The interpretation of the results of the gate analysis gives understandable information on the method, in which the model balances the contribution of local and global features in each sample and per dimension of features. We hope that dynamic feature fusion strategies provide an interesting future direction of hybrid medical image analysis and release the full model code to facilitate reproducibility and future studies.

\section{Acknowledgement}
The authors express their sincere gratitude to Hamdard University for providing the necessary support and resources to carry out this research. We also extend our appreciation to all co-authors for their valuable contributions throughout the study. Additionally, we acknowledge the use of ChatGPT for assistance in writing and grammar refinement.

\end{document}